%% file: large_scale_dnn.tex
\def\year{2017}\relax
\documentclass[letterpaper]{article}
\usepackage{aaai17}
\usepackage{times}
\usepackage{helvet}
\usepackage{courier}
\usepackage{amssymb}
\usepackage{algorithm}
\usepackage{algorithmic}
\usepackage{natbib}
\usepackage{textcomp}
\usepackage{amsmath}

\usepackage{enumitem}

\usepackage[draft]{todonotes} 

\usepackage[utf8]{inputenc} 
\usepackage[T1]{fontenc}    
\usepackage{url}            
\usepackage{booktabs}       
\usepackage{amsfonts}       
\usepackage{nicefrac}       
\usepackage{microtype}      
\usepackage{lipsum}
\include{everything}
\usepackage{wrapfig}
\usepackage{epstopdf}
\graphicspath{{./images/}}

\begin{document}

 \title{Distributed Hessian-Free Optimization for Deep Neural Network}
\author{
Xi He\\
 Industrial and Systems Engineering\\
Lehigh University, USA\\
  \texttt{xih314@lehigh.edu} \\
  \And
  Dheevatsa Mudigere \\
  Parallel Computing Lab \\
    Intel Labs, India \\
  \texttt{dheevatsa.mudigere@intel.com} \\
  \AND
    Mikhail Smelyanskiy \\
    Parallel Computing Lab \\
    Intel Labs, SC \\
    \texttt{mikhail.smelyanskiy@intel.com} \\
    \And
  Martin Tak\'a\v{c} \\
   Industrial and Systems Engineering \\
  Lehigh University, USA \\
  \texttt{takac.mt@gmail.com} \\
}


\maketitle
\begin{abstract}
Training deep neural network is a high dimensional and a highly non-convex optimization problem. In this paper, we revisit Hessian-free optimization method for deep networks with negative curvature direction detection. We also develop its distributed variant and demonstrate superior scaling potential to SGD, which allows more efficiently utilizing larger computing resources thus enabling large models and faster time to obtain desired solution. We show that these techniques accelerate the training process for both the standard MNIST dataset and also the TIMIT speech recognition problem, demonstrating robust performance with upto an order of magnitude larger batch sizes. This increased scaling potential is illustrated with near linear speed-up on upto 32 CPU nodes for a simple 4-layer network.


   \end{abstract}

\section{Introduction}
Deep learning has shown great success in many practical applications, such as image classification \cite{krizhevsky2012imagenet,simonyan2014very,heResNet}, speech recognition \cite{hinton2012deep,seide2014parallelizability,deepspeech2}, etc. Stochastic gradient descent (SGD), as one of the most well-developed method for training neural network, has been widely used. Besides, there has been plenty of interests in second-order methods for training deep networks \cite{martens2010deep}. The reasons behind these interests are multi-fold. At first, it is generally more substantial to apply weight updates derived from second-order methods in terms of optimization aspect, meanwhile, it takes roughly the same time to obtain curvature-vector products \cite{kiros2013training} as it takes to compute gradient which make it possible to use second-order method on large scale model. Furthermore, computing gradient and curvature information on large batch (even whole dataset) can be easily distributed across several nodes. Recent work has also been used to reveal the significance of identifying and escaping saddle point by second-order method, which helps prevent the dramatic deceleration of training speed around the saddle point \cite{dauphin2014identifying}.

Line search Newton-CG method (also known as the truncated Newton Method), as one of the practical techniques to achieve second-order method on high dimensional optimization, has been studied for decades \cite{nocedal2006numerical}. Note that Newton-CG method does not require explicit knowledge of Hessian matrix, and it requires only the Hessian-vector product for any given vector. One special case for using Hessian-vector product is to train deep neural network, also known as Hessian-free optimization, and such Hessian-free optimization is exactly used in Marten's HF \cite{martens2010deep} methods.

It is well known that traditional SGD method is inherently sequential and becomes very expensive (time-to-train) to apply on very large data sets. More detail discussion can be found in \cite{zhang2016distributed}, wherein Momentum SGD (MSGD) \cite{sutskever2013importance}, ASGD and MVASGD \cite{polyak1992acceleration}, are considered as alternatives. However, it is shown that these methods have limited scaling potential, due to the limited concurrency. However, unlike SGD, Hessian-free method can be distributed naturally, allow for large mini-batch sizes (increased parallelism) while improving convergence rate and also the better the quality of solution - we are therefore motivated to develop a distributed variant of Hessian-free optimization.
\begin{figure*}
\centering
\includegraphics[width=4cm]{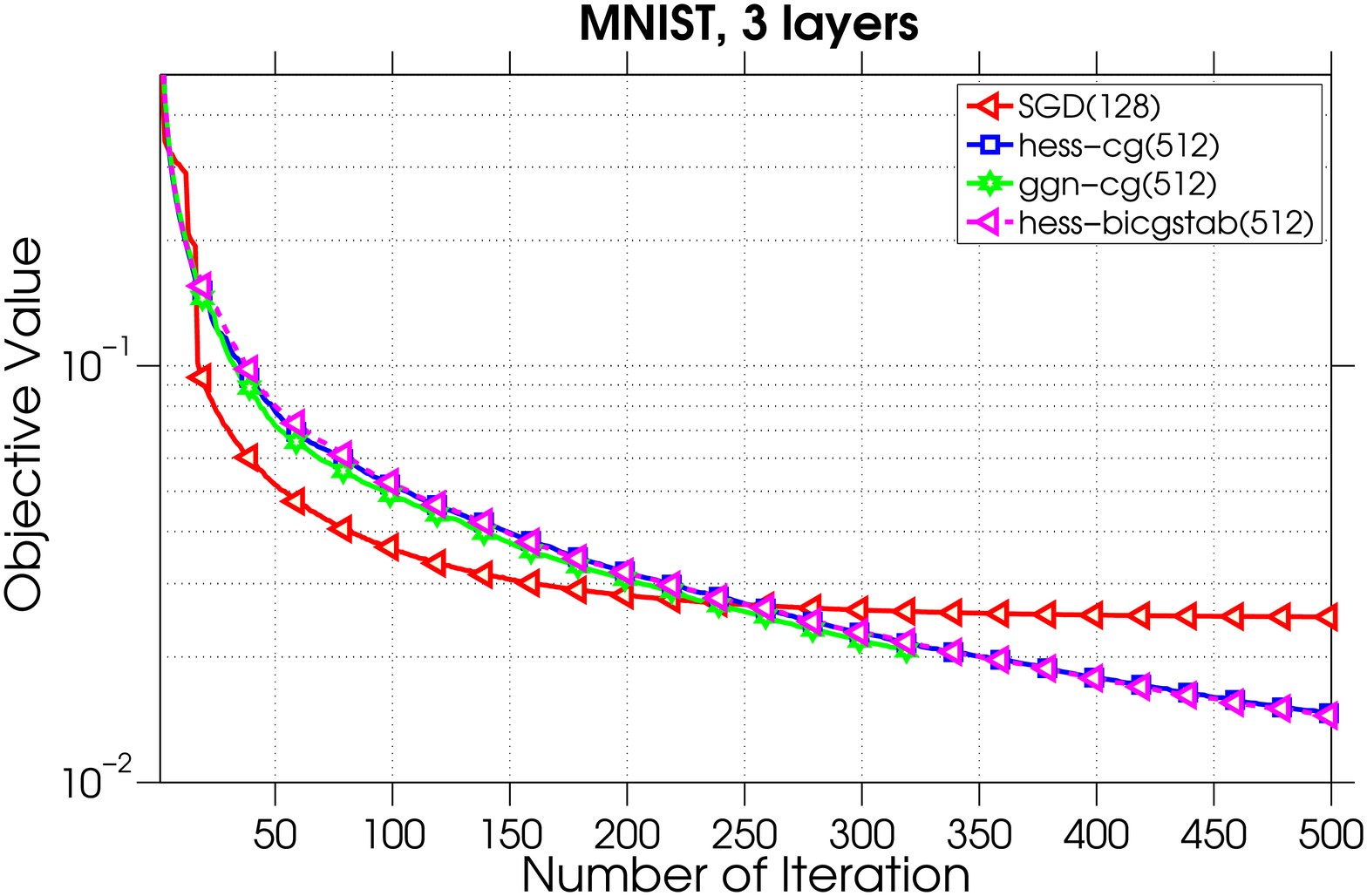}
\includegraphics[width=4cm]{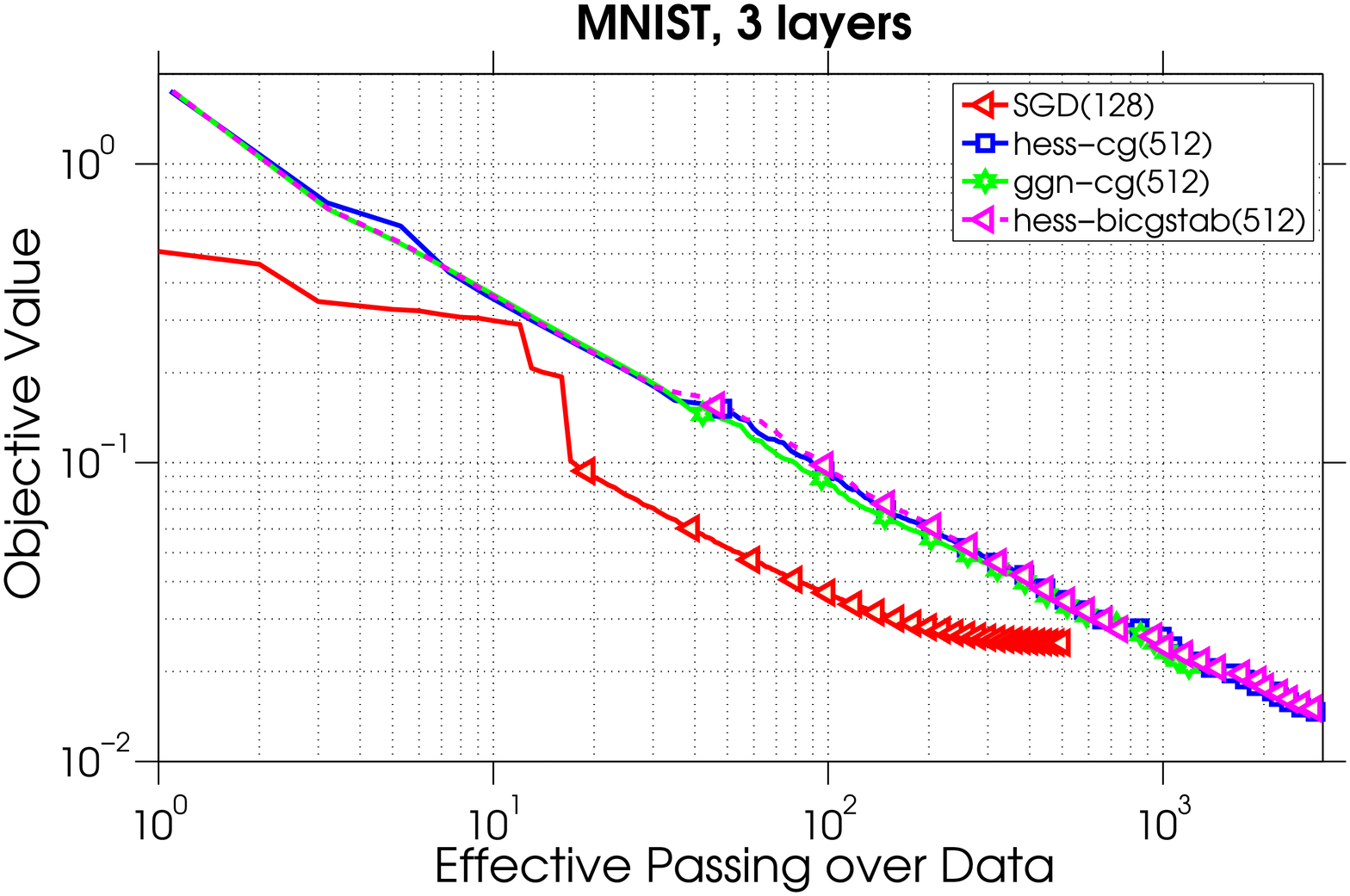}
\includegraphics[width=4cm]{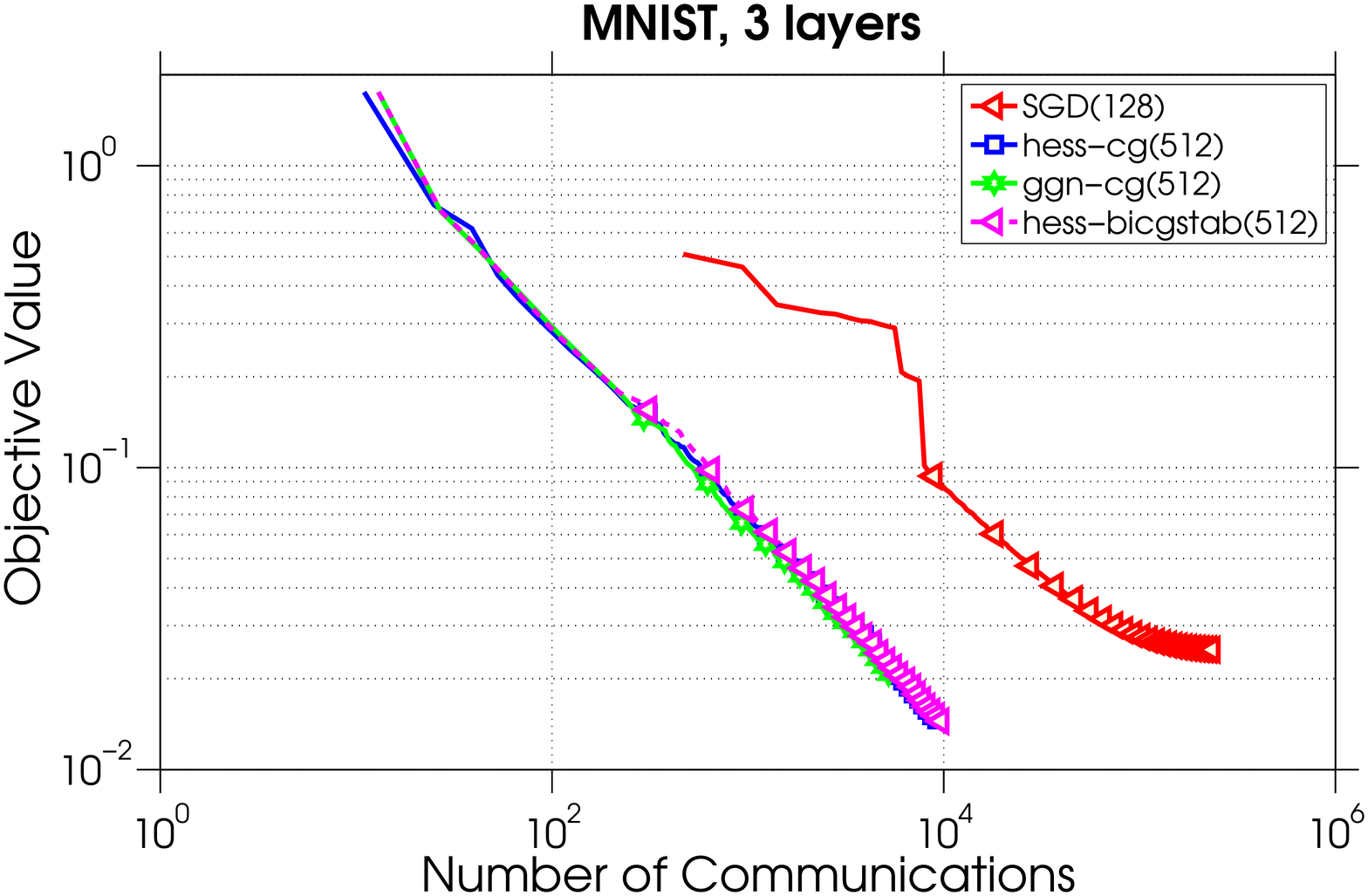}
\caption{Performance comparison among SGD and Hessian-free variants.}
\label{fig: comparison_sgd_hf_performance}
\end{figure*}

In this paper, we explore the Hessian-free methods to develop more robust and scalable solver for deep learning. We discuss novel ways to utilize negative curvature information to accelerate training speed. This is different with original Marten's HF, where the negative curvature is ignored by either using Gauss-newton Hessian approximation or truncated Newton method. We perform experimental evaluations on two datasets without distortions or pre-training: hand written digits recognition (MNIST) and speech recognition (TIMIT).

Additionally, we explore Hessian-free methods in a distributed context. Its potential scaling property is discussed, showcasing scaling potential of distributed Hessian-free method and how it allows taking advantage of more computing resources without being limited by the expensive communication.

{\bf Contributions.}
\begin{itemize}[leftmargin=0.5cm,topsep=0pt,itemsep=0ex,partopsep=0ex,parsep=0ex]

\item In this paper, we propose an algorithm which outperforms Newton-CG method.
This is achieved by considering
negative curvature information. The algorithm is able to escape saddle points in a cheaper manner and therefore have better training performance.

\item  We evaluate the distributed variant of this second-order method, showcasing its superior scaling property compared to conventional SGD.
\item We compare and analyze different methods both from algorithmic (convergence) and computing perspectives. We show in this paper that by using distributed Hessian-free method, we are able to achieve much better and stable scaling performance in terms of nodes and size of mini-batch.
\end{itemize}



\section{Distributed Hessian-free Optimization Algorithms} 
\label{sec:distributed_Hessian_free_algorithm}

DNN training can be parallelized using the following two strategies - model parallelism (we split weights across many computing nodes) and data parallelism (when the data is partitioned across nodes).






{\bf Model Parallelism.}
In the model parallelism the weights of network are split across $N$ nodes. In one SGD iteration all nodes work on the same data but each is responsible only for some of the features.
Hence after each layer they have to synchronize to have the activations needed for the portion of the model they have for in next layer. For the backward pass they have to also synchronize after each layer and exchange the $\delta$'s used to compute gradients. After gradients are computed they can be applied to weights stored locally.

If a mini-batch  of size $b$ is used and the weights for hidden layer have dimensions $d_1 \times d_2$, then each node (if split equally) will have to store  $\frac{d_1\times d_2}{N}$ floats. The total amount of data exchanged over network for this single layer is $d_1 \times b$. If we consider a deeper network with dimensions $d_1, d_2, \dots, d_l$ then the total number of floats to be exchanged in one epoch of SGD is approximately $2\times \frac{n}{b} \times b \sum_i d_i$ and total number of communications (synchronizations) needed per one epoch is $2\times l \times \frac{n}{b}$.


{\bf Data Parallelism.}
The other natural way how to implement distributed SGD for DNN is to make a copy of weights on each node and split the data across $N$ nodes, where each node owns roughly $n/N$ samples.
When a batch of size $b$ is chosen, on each node only $\frac{b}{N}$ samples are propagated using forward and backward pass. Then the gradients are reduced and applied to update  weights. We then have to make sure that after each iteration of SGD all weights are again synchronized. In terms of the amount of data sent over the network, in each iteration of SGD we have to reduce the gradients and broadcast them back. Hence amount of data to be send over the network in one epoch is $\frac{n}{b} \times  \log(N) \times \sum_{i=1}^{l} d_0 \times d_i$, where $d_0=d$ is the dimension of the input samples. Total number of MPI calls per epoch is hence only $\frac{n}{b}\times 2$ which is considerably smaller then for the model parallelism approach.

{\bf Limits of SGD.}
As it can be seen from the estimates for amount of communication and the frequency of communication, choosing large value of $b$ will minimize communication and for data parallelism also amount of data sent. However, as it was observed e.g. in \cite{takavc2013mini} that SGD (even for convex problem) can benefit from mini-batch only for small batch size $b$. After increasing $b$ above a critical value $\tilde b$, number of iterations needed to achieve a desired accuracy will not be decreased much if batch size $b>\tilde b$. Quite naturally this can be observed also for training DNN \cite{das2016distributed,zhang2016distributed}.

{\bf Benefits of Distributed HF. } As we will show in following sections, distributed HF needs less synchronizations/communications per epoch. SGD requires synchronization after each update (mini-batch). In distributed HF, one needs to synchronize only once for gradient computation and
then several times when solving the Newton system
e.g. using Conjugate Gradient (CG) method.




Let us in next Section describe the distributed Hessian-free algorithm. We assume that the size of the model is not huge and hence we choose data parallelism  paradigm. We assume that the samples are split equally across $K$ computing nodes (MPI processes).


\subsection{Distributed HF Optimization Framework}
Within this Hessian-free optimization approach, for the sake of completeness, we first state the general Hessian-free optimization method \cite{martens2010deep} in Algorithm~\ref{Alg: HF}.
\begin{algorithm}
    \caption{The Hessian-free optimization method}
    \label{Alg: HF}
    \begin{algorithmic}[1]
        \FOR {$k = 1, 2, \dots $}
            \STATE $g_k = \nabla f(\theta_k)$
            \STATE Compute/adjust damping parameter $\lambda$
            \STATE Define $B_k(d) = H(\theta_k)d + \lambda d$
            \STATE $p_k = \mbox{CG-Minimize}(B_k, -g_k)$
            \STATE $\theta_{k+1} = \theta_k + p_k$
        \ENDFOR
    \end{algorithmic}
\end{algorithm}
Here $\theta\in \R^N$ is the parameters of this neural network. At $k$-th iteration, full gradient of error function $f(\theta_k)$ is evaluated and (approximated) Hessian matrix is defined as $H(\theta_k)$. Based on this (approximated) Hessian and a proper damping parameter, which aims to make the damped Hessian matrix $B_k$ positive definite and/or avoid $B_k$ being singular. Following this, a quadratic approximation of $f$ around $\theta_k$ is constructed as

\begin{equation}
    m_k(d ) :=
    f(\theta_k)
    + g_k^Td + \frac{1}{2}d^TB_kd.
\end{equation}
If $B_k$ is positive definite, then we can obtain Newton step $d_k$ by letting $d_k := \arg\min_d m(d) = -B_k^{-1}g_k$. Otherwise, we solve $\min_d m(d)$ by CG method and choose the current iteration whenever a negative curvature direction is encountered, i.e., exist a conjugate direction $p$, such that $p^TB_kp < 0 $. If the negative curvature direction is detected at the very first CG iteration, the steepest descent direction $-g_k$ is selected as a descent direction.

 \begin{figure*}
\centering
\includegraphics[width=4cm]{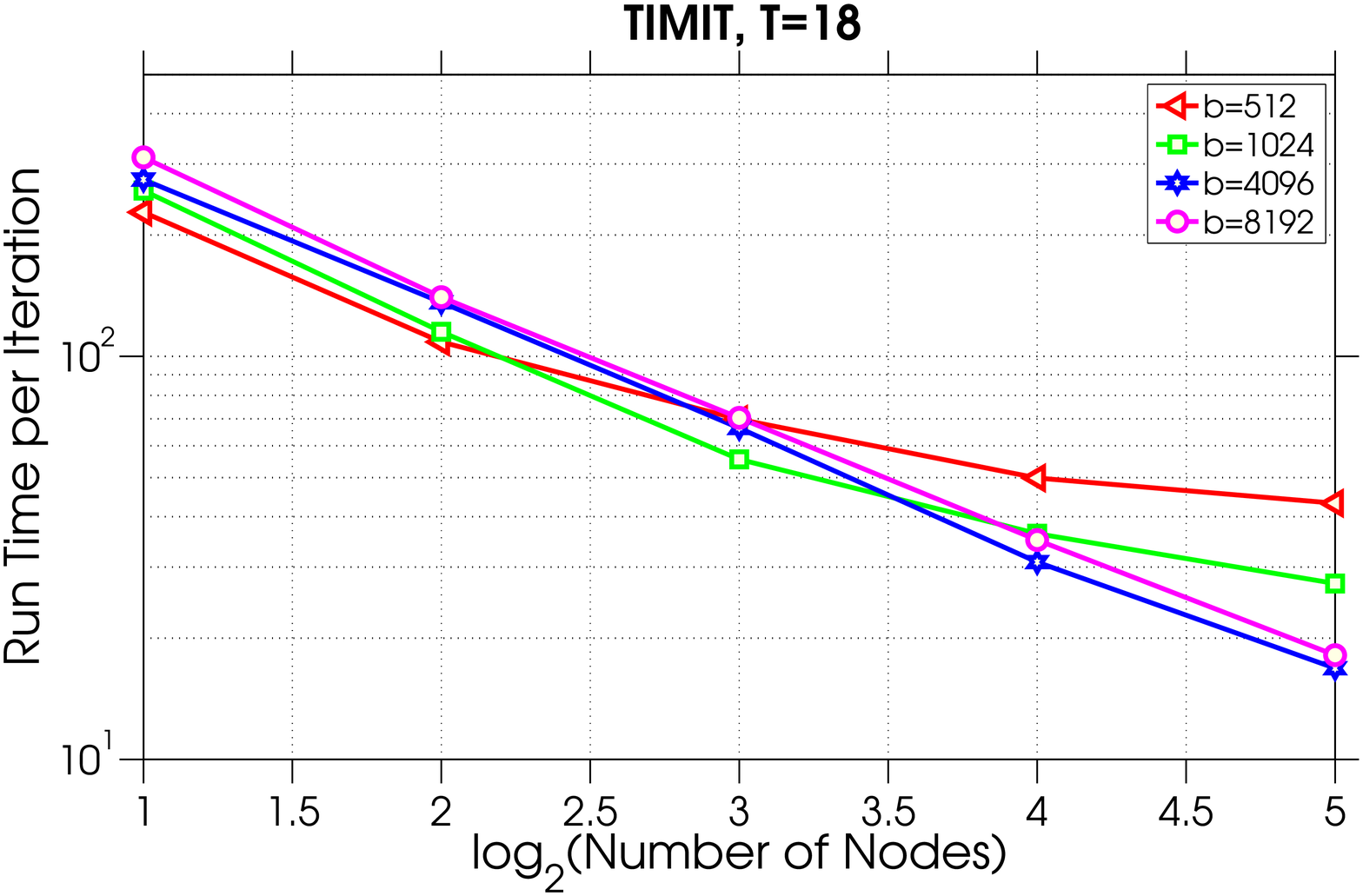}
\includegraphics[width=4cm]{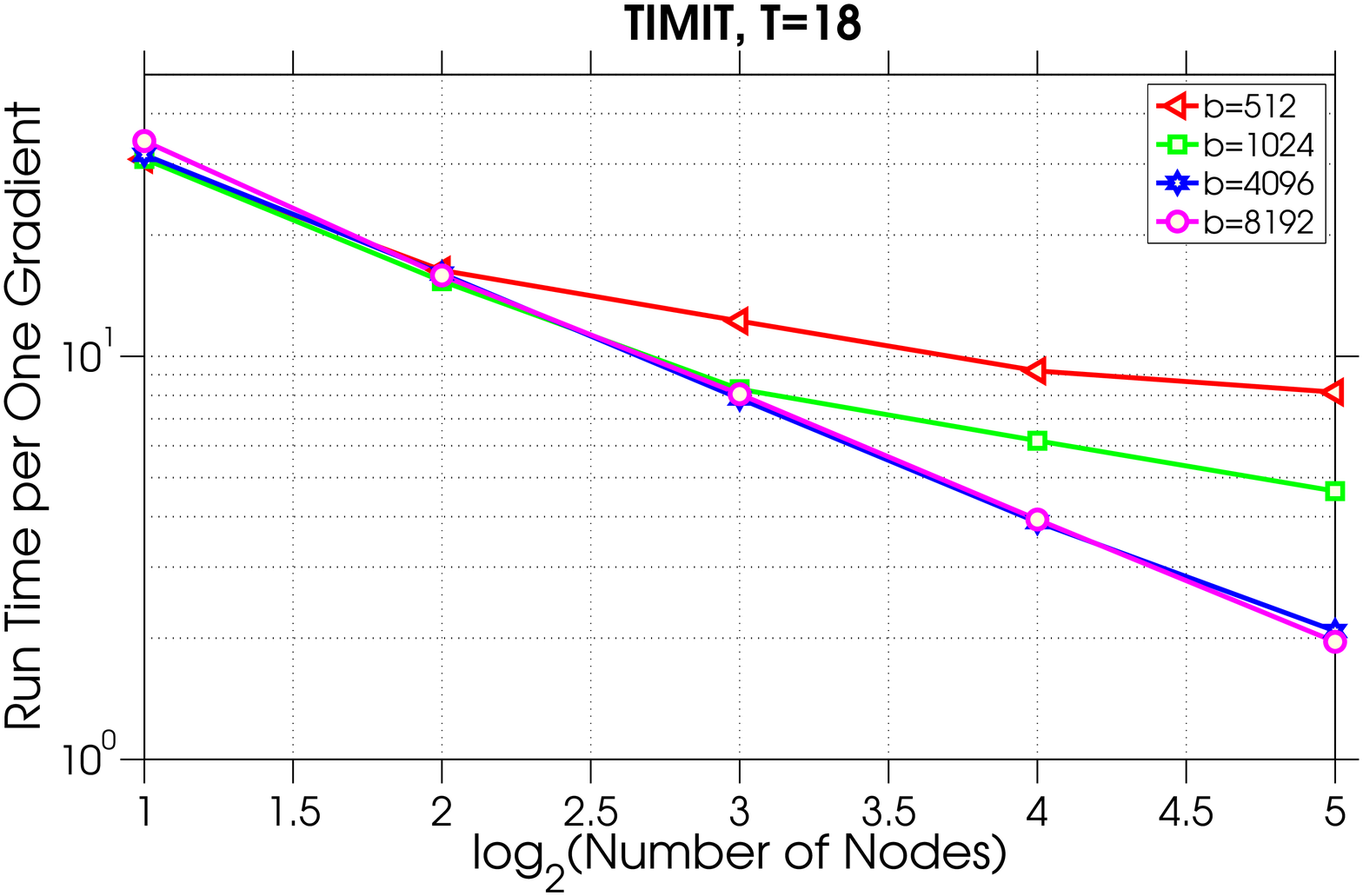}
\includegraphics[width=4cm]{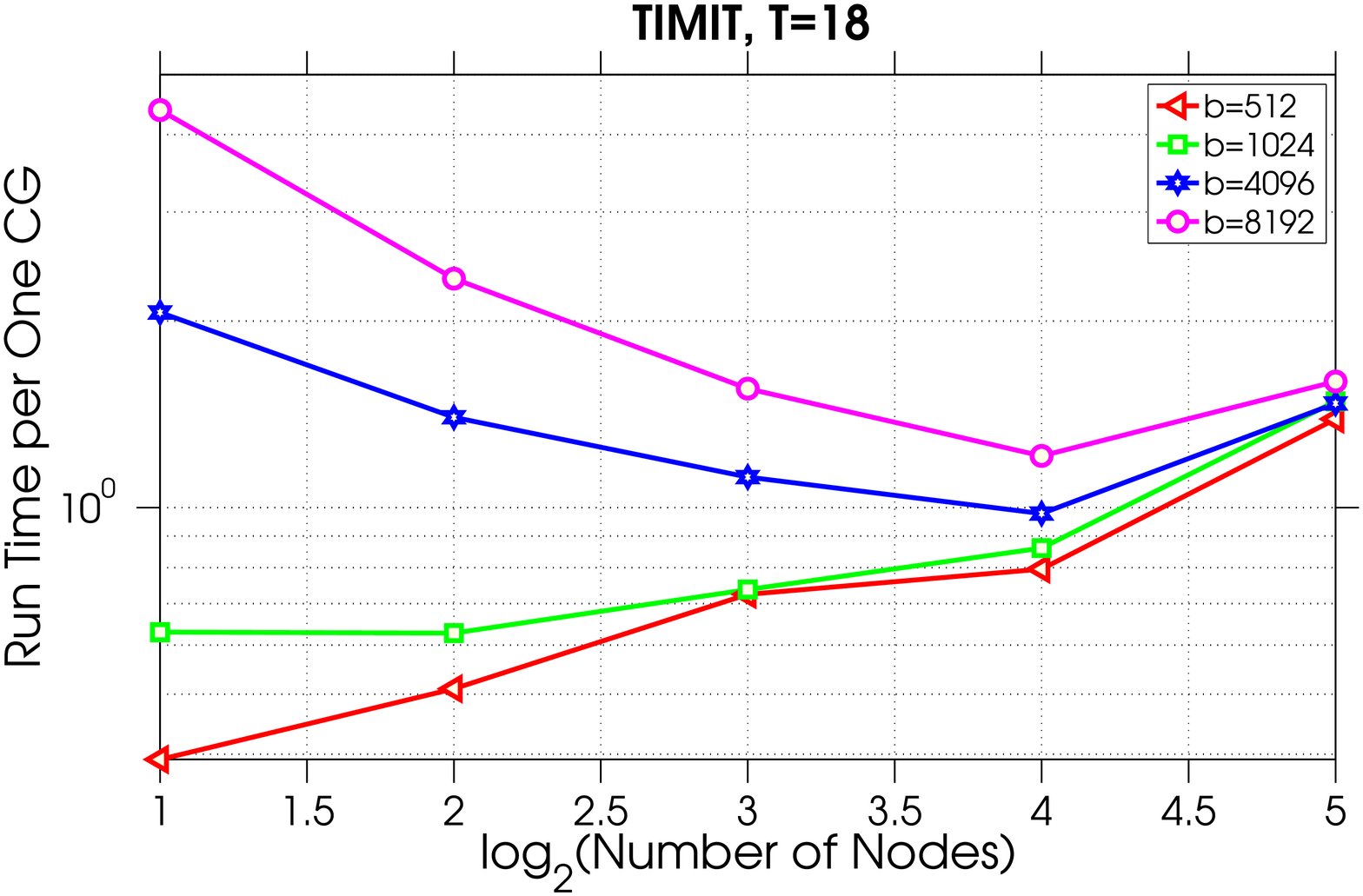}
\includegraphics[width=4cm]{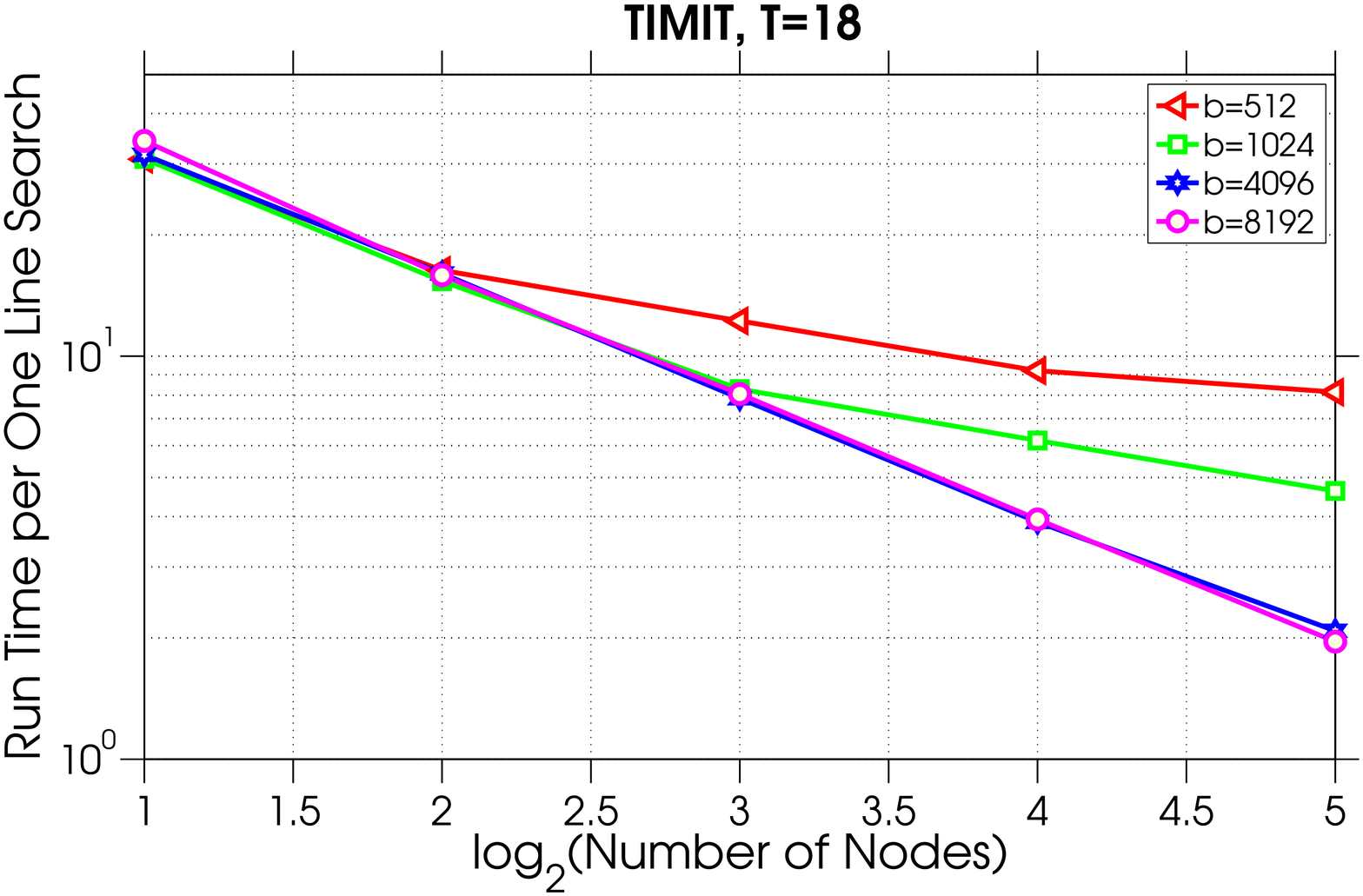}

\includegraphics[width=4cm]{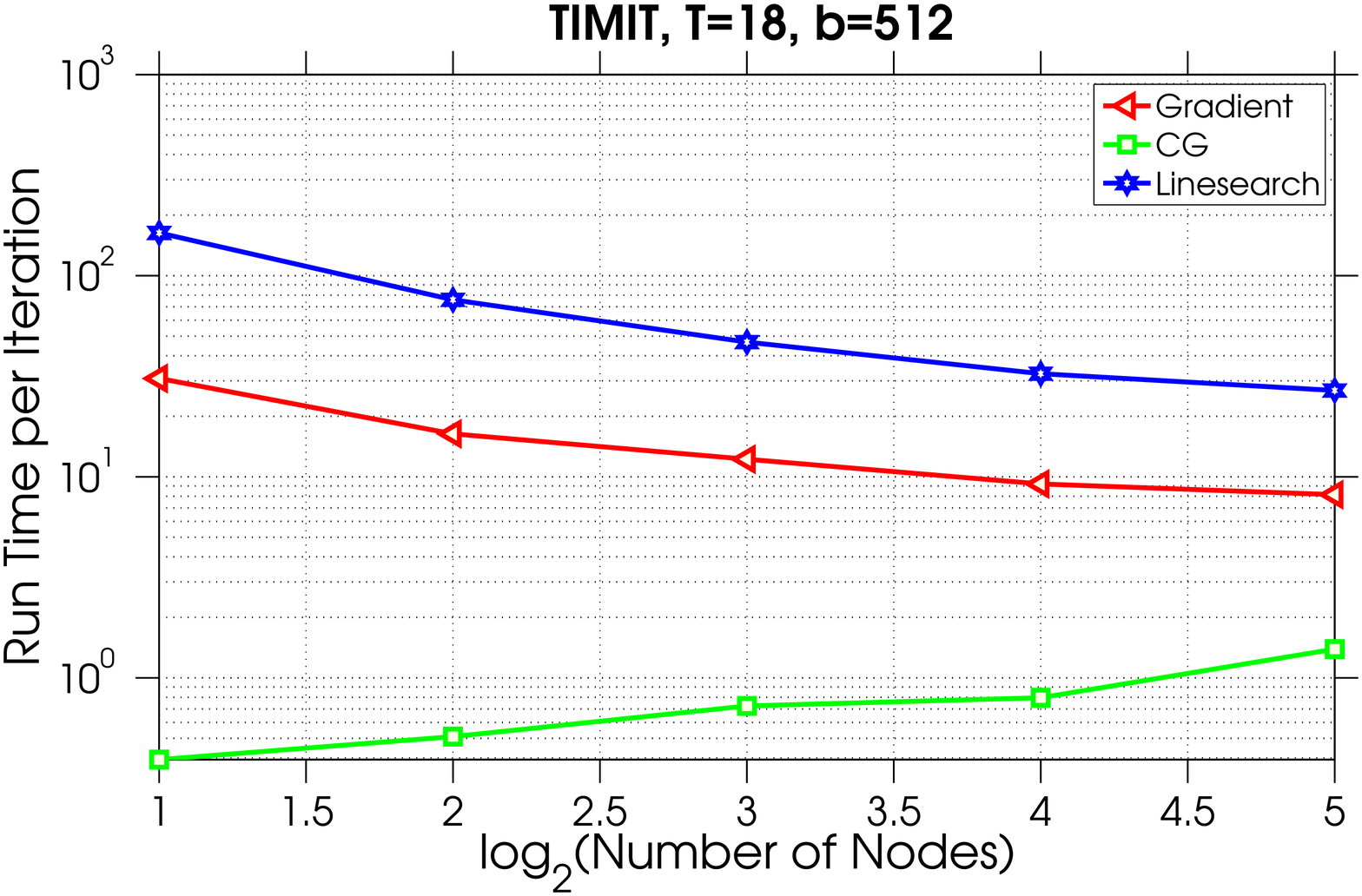}
\includegraphics[width=4cm]{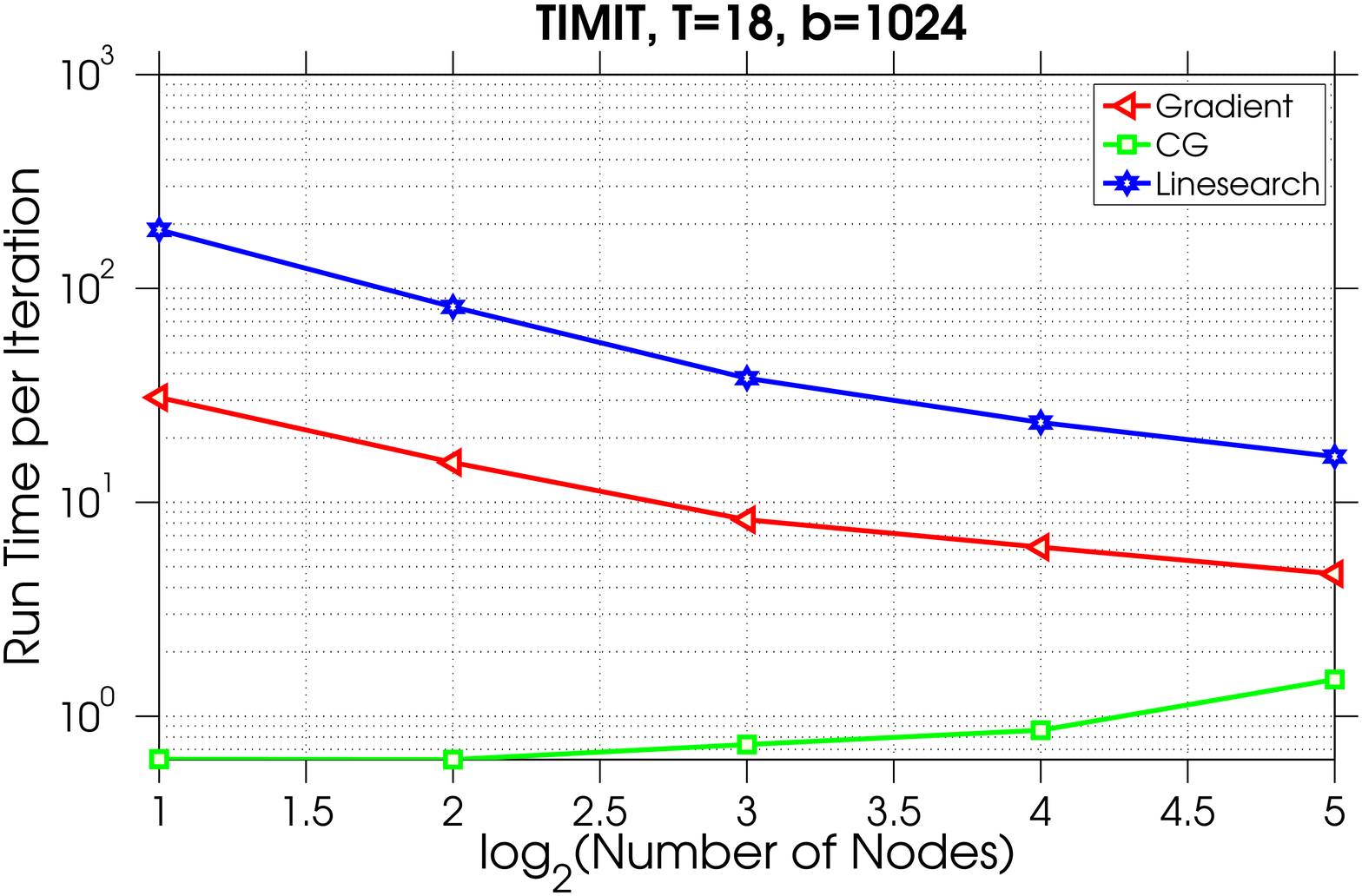}
\includegraphics[width=4cm]{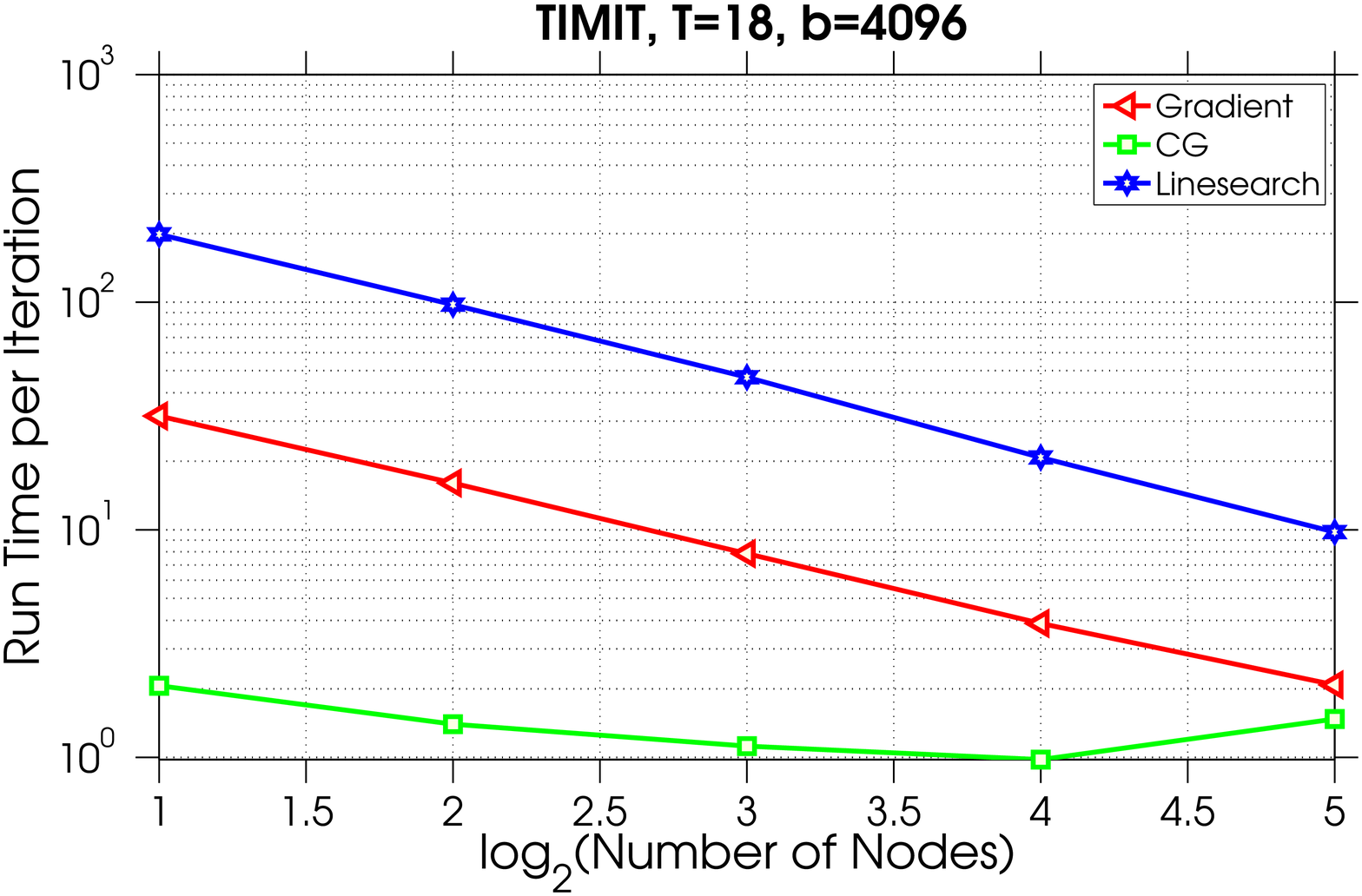}
\includegraphics[width=4cm]{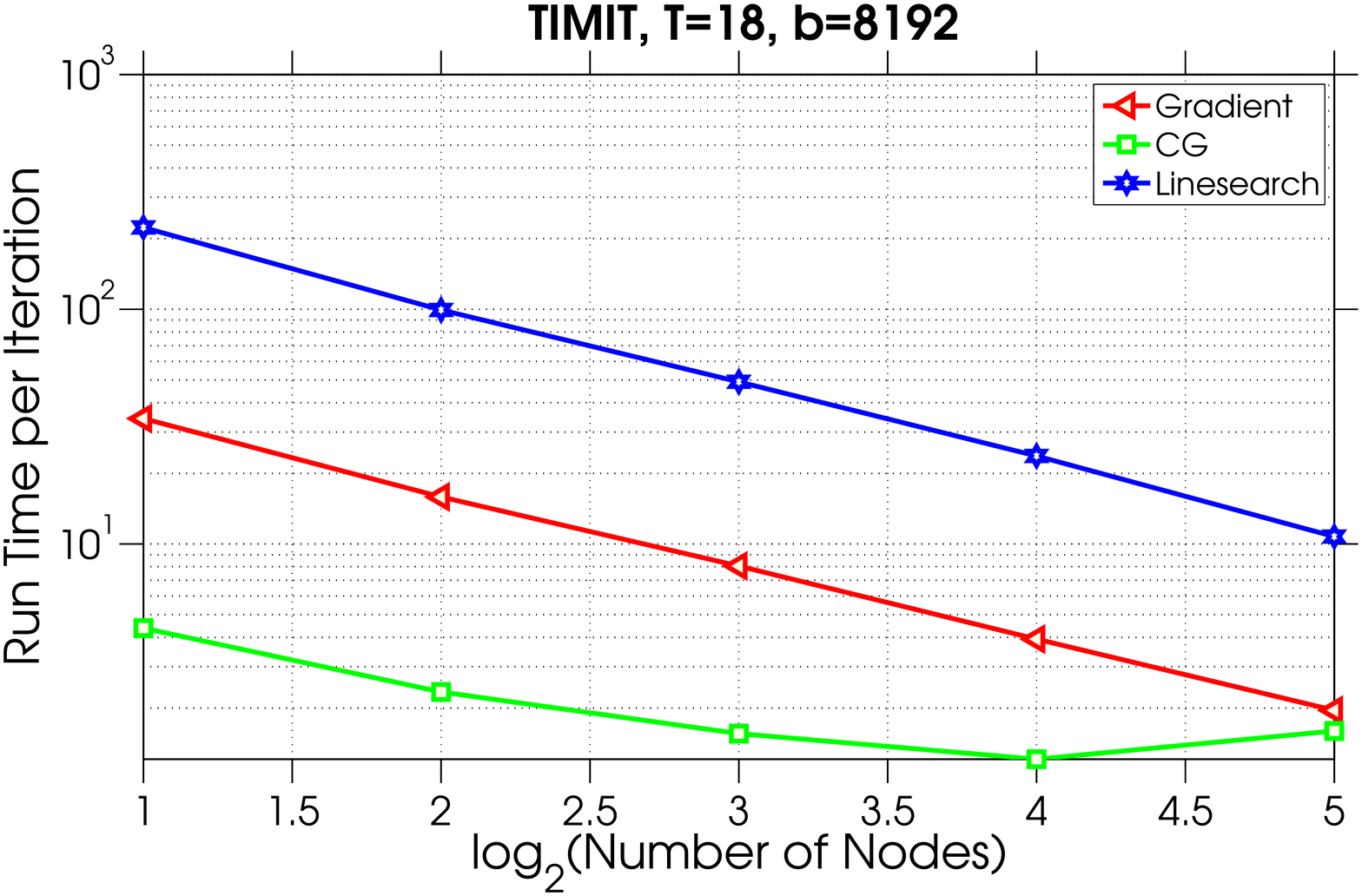}
\caption{Performance scaling of different part in distributed HF on upto 32 nodes (1,152 cores).}
\label{fig: Timit_all_appendix}
\end{figure*}

\cite{martens2010deep} modified Algorithm \ref{Alg: HF} in several ways to make it suitable for DNNs. Within neural network, Hessian-vector can be calculated by a forward-backward pass which is roughly twice the cost of a gradient evaluation by using $\mc{R}$-operator. On the other side, due to non-convexity of error function $f$, Hessian matrix is more likely to be indefinite and therefore a Gauss-Newton approximated Hessian-matrix is used. Note that Gauss-Newton is positive semi-definite matrix but it can be treated as a good approximation when the current point is close to local minimizer, which also motivates our work to design a Hybrid approach. Moreover,   pre-conditioning and a CG-backtracking technique is used to decrease the number of CG iterations and obtain the \textit{best} descent direction.
However, it is claimed in \cite{wiesler2013investigations} that such techniques are not very helpful and even can lead to degraded performance, increased computing and storage requirements. Therefore, we skip these steps and directly move on to our distributed HF algorithm depicted in Algorithm \ref{Alg: dis_HF}.
\begin{algorithm}
    \caption{Distributed Hessian-Free Algorithm}
    \label{Alg: dis_HF}
    \begin{algorithmic}[1]
    \footnotesize
        \STATE {\bf Initialization:}
        $\theta_0$ (initial weights), $\lambda$ (initial damping parameter), $\delta_{0}$ (starting point for CG solver), $N$ (number of MPI processes), distributed  data         \FOR {$k = 1, 2, \dots $}
            \STATE Calculate gradient $\nabla f_{[i]}(\theta_k)$ on each node $i = 0, \dots, N-1$
            \STATE Reduce $\nabla f_{[i]}(\theta_k)$ to root node to obtain full gradient $g_k = \frac{1}{N}\sum_{i=0}^{N-1}\nabla f_{[i]}(\theta_k)$
            \STATE Construct stochastic (approximated) Hessian-vector product operator $G_k(v)$
            \begin{itemize}
                \item  Calculate Hessian-vector product $\nabla^2 f_{[i]}(\theta_k)v$ corresponding to one Mini-batch on each node $i = 0, \dots, N-1$
                \item  Reduce $\nabla^2 f_{[i]}(\theta_k)v$ to root node to obtain $G_k(v) = \frac{1}{N}\sum_{i=0}^{N-1}\nabla^2 f_{[i]}(\theta_k)v$
            \end{itemize}
            \STATE Solve $G_k(v) = -g_k$ by BI-CG-STAB method with starting point $0$ or $\eta\delta_{k-1}$ ($\eta$ is decay)

            \STATE Use CG solution $s_k$ or possible negative curvature direction $d_k$ to find the best descent direction $\delta_k$


            \STATE Find $\alpha_k$ satisfying $f(\theta_k+\alpha_k \delta_k) \leq f(\theta_k) + c\alpha_k g_k^T\delta_k$ ($c$ is a parameter)

            \STATE \textbf{Update} $\theta_{k+1} = \theta_k+\alpha_k \delta_k$
        \ENDFOR
    \end{algorithmic}
\end{algorithm}
 For example,   to calculate full gradient (or Hessian vector product needed by BI-CG-STAB \footnote{BI-CG-STAB is a variant of CG which is used to solve indefinite system. The numerical results in this paper are obtained  using Bi-CG-STAB.} solver), each node is responsible for computing the gradient (and Hessian vector products) based on data  samples
stored locally. A reduction step is followed to aggregate them to a root node.
\subsection{Dealing with Negative Curvature}
As mentioned in \cite{dauphin2014identifying}, to minimize a non-convex error functions over continuous, high dimensional spaces, one may encounter proliferation of saddle points which are surrounded by high error plateaus.  One shortage coming from the use of  first-order methods like SGD is that it can not recognize curvature information, and therefore dramatically slow down the learning rate around such saddle points. The saddle-free Newton method (SFN) \cite{dauphin2014identifying} is then proposed to identify and escape such saddle points. To achieve this, they build an exact Hessian to accomplish SFN on a small size neural network. However, this is impractical or even infeasible for medium or large scaled problems.  In this paper, we propose another method to exploit the local non-convexity of the error function even for a large size network.

A negative curvature direction at current point $\theta$ of function $f$ is defined as a vector $d\in \R^n/\{0\}$, such that it is dominant in the negative eigenspace ($d^THd <0$), where $g, H$ are gradient and Hessian of $f$ at point $\theta$. By letting $\tilde{d} = -\mbox{sgn}(g^Td)d$, where $\mbox{sgn}(x) = 1, x\geq0; \mbox{sgn}(x)=-1$, we are able to always find a descent direction, since $g^T\tilde{d} < 0$.

Actually, along with those negative directions, the approximated quadratic model is unbounded below, which shows potential of reduction at such direction (at least locally, while the quadratic approximation is valid).
 It was  shown in \cite{olivares2008nonconvex} that if algorithms uses negative curvature directions, it will  eventually converge to second-order critical point.

We are now ready to show an improved method to find a possible negative curvature by stabilized bi-conjugate gradient descent (Bi-CG-STAB, Algorithm \ref{Alg: bicg_stab}), which is a Krylov method that can be used to solve unsymmetrical or indefinite linear system \cite{saad2003iterative}.
The benefits of using Bi-CG-STAB is that
we can  use exact stochastic Hessian information (which may not be positive definite) instead of using Gauss-newton approximation, since the later one will lose the curvature information.
It is shown in \cite{martens2010deep} that HF-CG is unstable and usually fails to convergence.
The reason behind that is a fact that HF-CG
ignores  negative curvature.
At the point where the Hessian has relative large amount of negative eigenvalues, it is also inefficient to find a descent direction by restarting the CG solver and modifying the damping parameter.

To use BI-CG-STAB, we set a fixed number of iterations \cite{kiros2013training} and choose the candidates of descent direction for CG-backtracking \cite{martens2010deep} by letting $\tilde{d} = -\mbox{sign}(g^Td)d$. Therefore, at each CG iteration, either a Newton-type descent direction where $\tilde{d}^TH\tilde{d} >0, g^T\tilde{d} <0$ is found or a negative curvature descent direction where $\tilde{d}^TH\tilde{d}<0, g^T\tilde{d}<0$ is found. By combining Amijo line search (see Algorithm \ref{Alg: dis_HF}), it is guarantee to have monotone decrease on the objective function value and the saddle point would be escaped whenever a negative curvature direction is detected around the saddle plateaus.

\begin{algorithm}
    \caption{Bi-CG-STAB Algorithm}
    \label{Alg: bicg_stab}
    \begin{algorithmic}[1]
        \STATE Compute $r_0 := b - Ax_0$. Choose $r_0^*$ such that $(r_0, r_0^*) \neq 0$
        \STATE $p_0 := r_0$, $k:=0$
        \IF {Termination condition not satisfied}
        \STATE $\alpha_j := (r_j, r_0^*) / (Ap_j, r_0^*)$
        \STATE $s_j := r_j - \alpha_jAp_j$
        \STATE $\gamma_j := (s_j, As_j) / (As_j, As_j)$
        \STATE $x_{j+1} := x_j + \alpha_jp_j + \gamma_j s_j$
        \STATE $r_{j+1} := s_j - \gamma_jAs_j$
        \STATE $\beta_j := \tfrac{(r_{j+1}, r_0^*)}{(r_j, r_0^*)} \times \tfrac{\alpha_j}{\gamma_j}$
        \STATE $p_{j+1}:=r_{j+1} + \beta_j(p_j-\gamma_jAp_j)$
        \ENDIF
    \end{algorithmic}
\end{algorithm}

\section{Numerical Experiments}
\label{sec: numerical experiments}

We study the multi-node scalability on the Endeavor cluster. Each Endeavor compute node has two Intel\textregistered Xeon\texttrademark{\sc E5-2697v4} processors ($18$x$2$ cores), at a clock speed of $2.3$ {\sc GHz} and $128$ {\sc GB DDR4} memory. We use Intel {\sc mpi} $5.1.3.181$, and Intel compiler {\sc icc} $16.0.2$.

\begin{figure*}
\centering
\includegraphics[width=4cm]{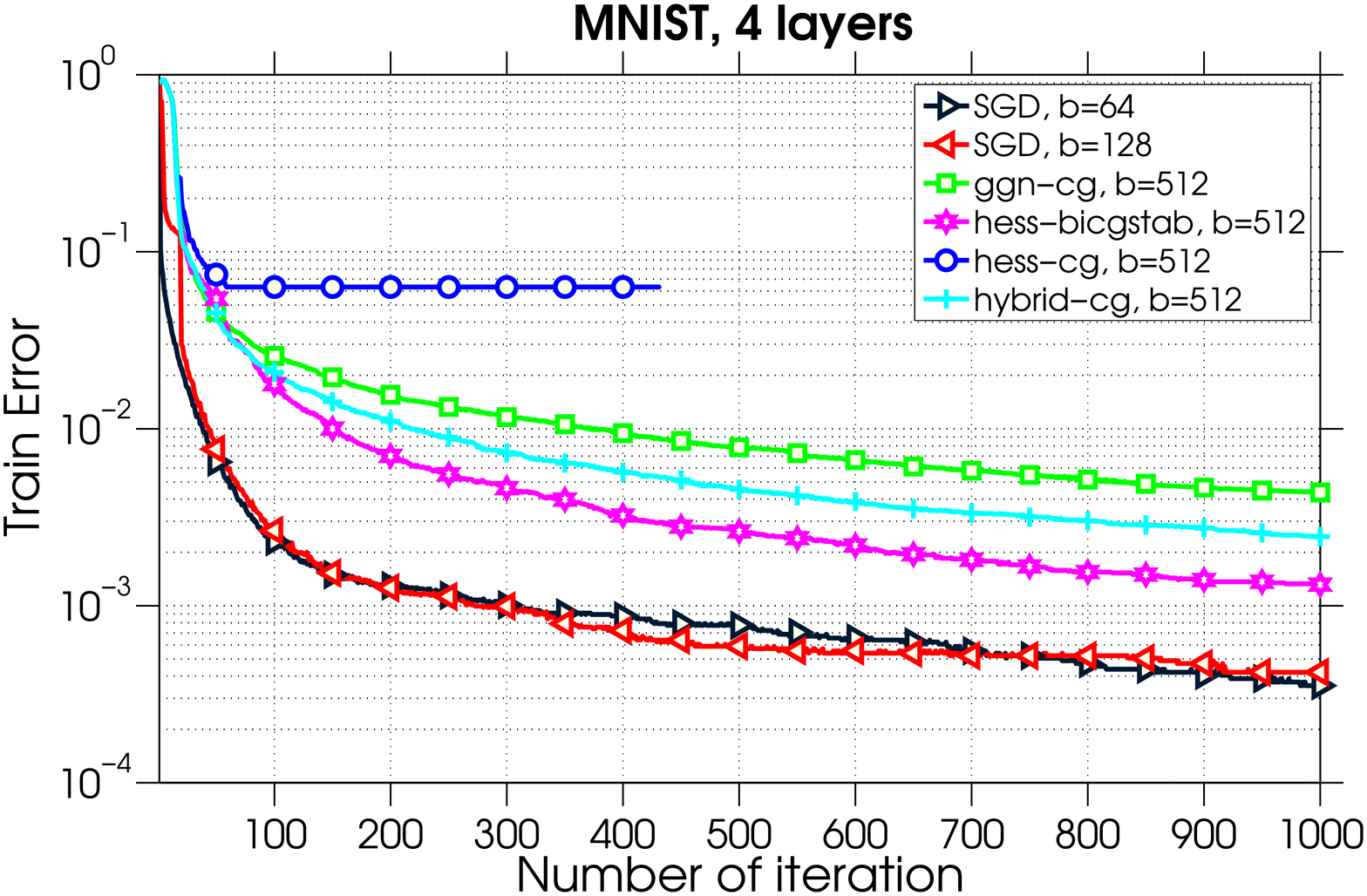}
\includegraphics[width=4cm]{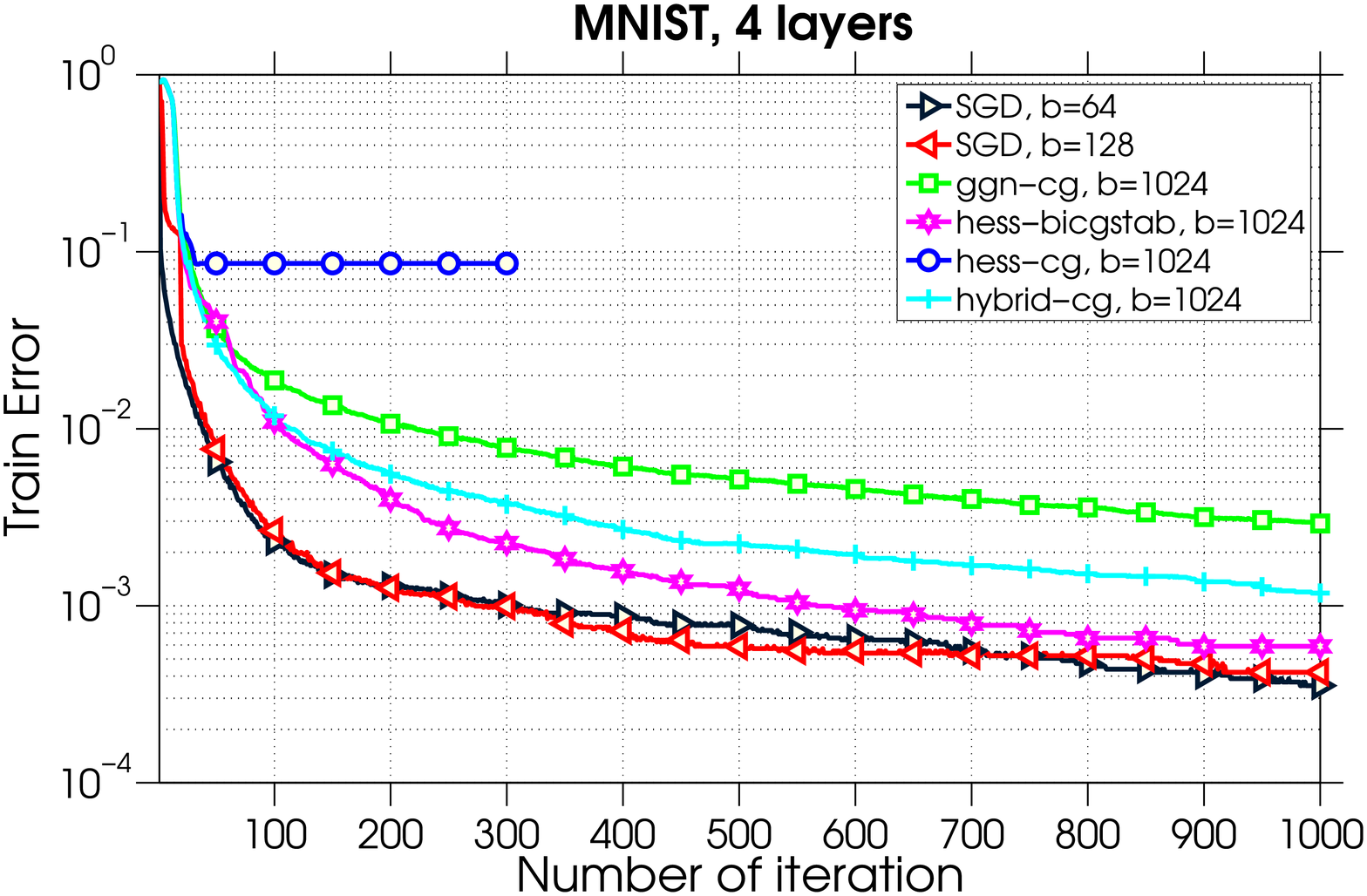}
\includegraphics[width=4cm]{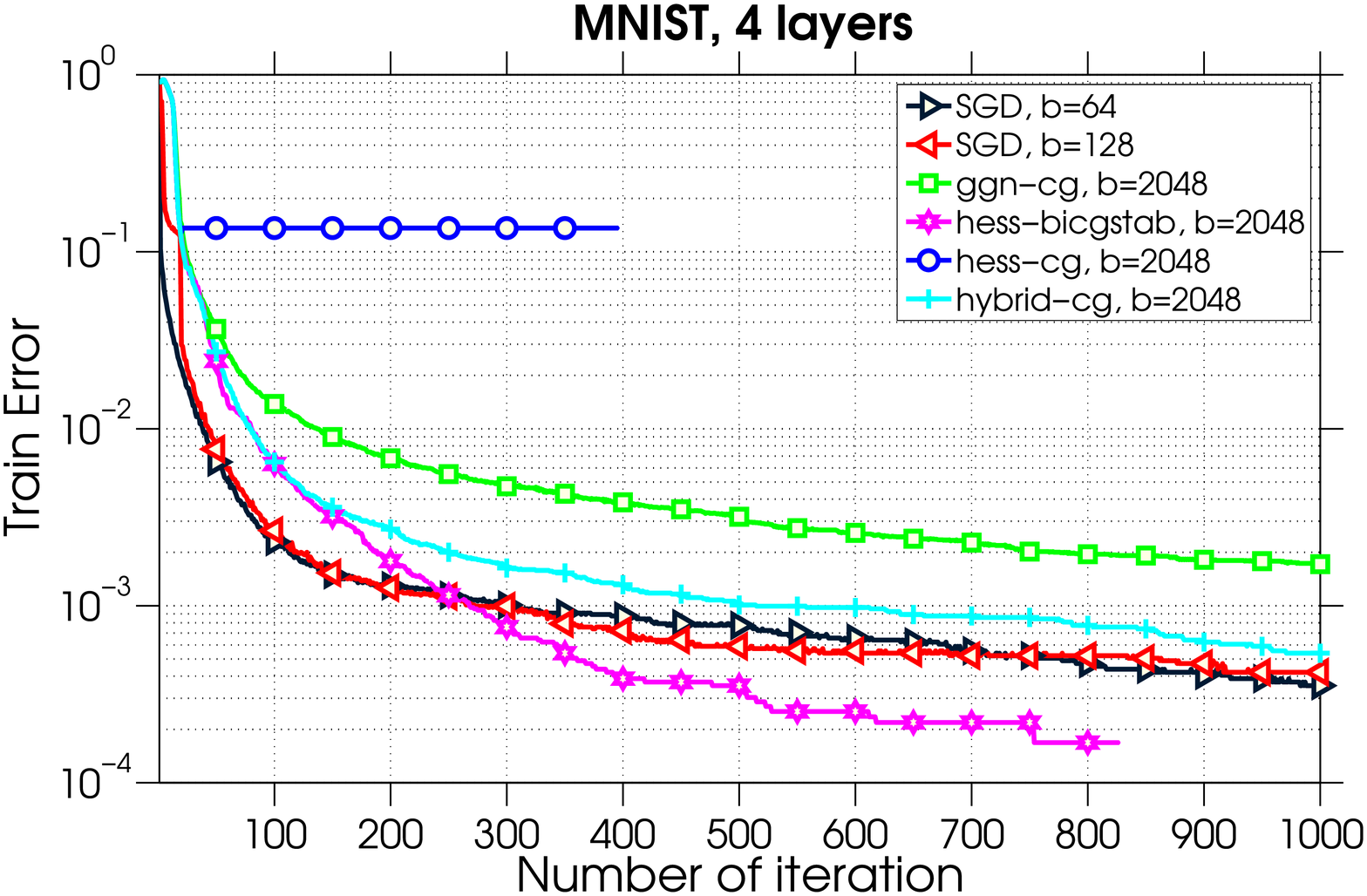}
\includegraphics[width=4cm]{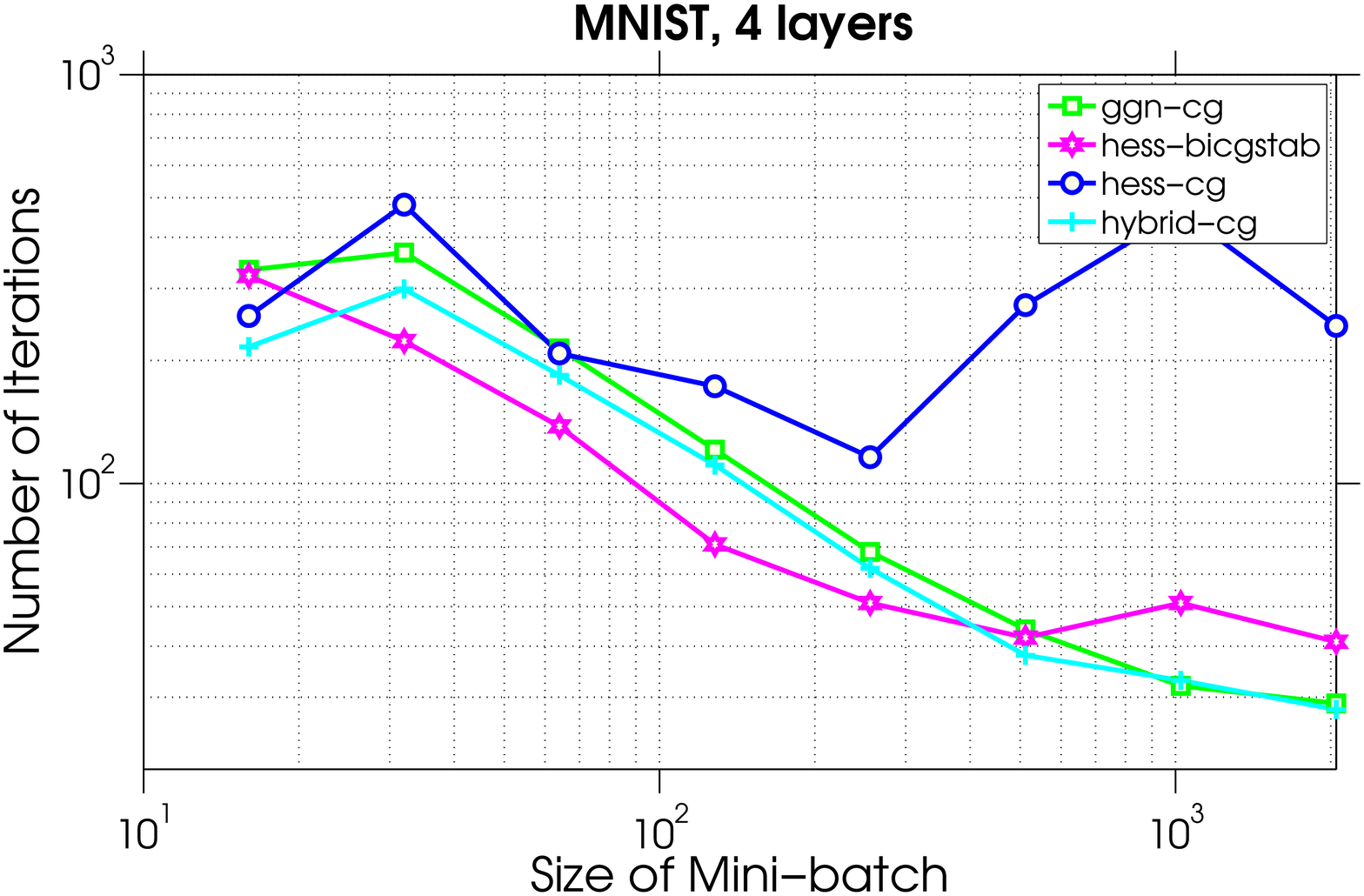}\\
\includegraphics[width=4cm]{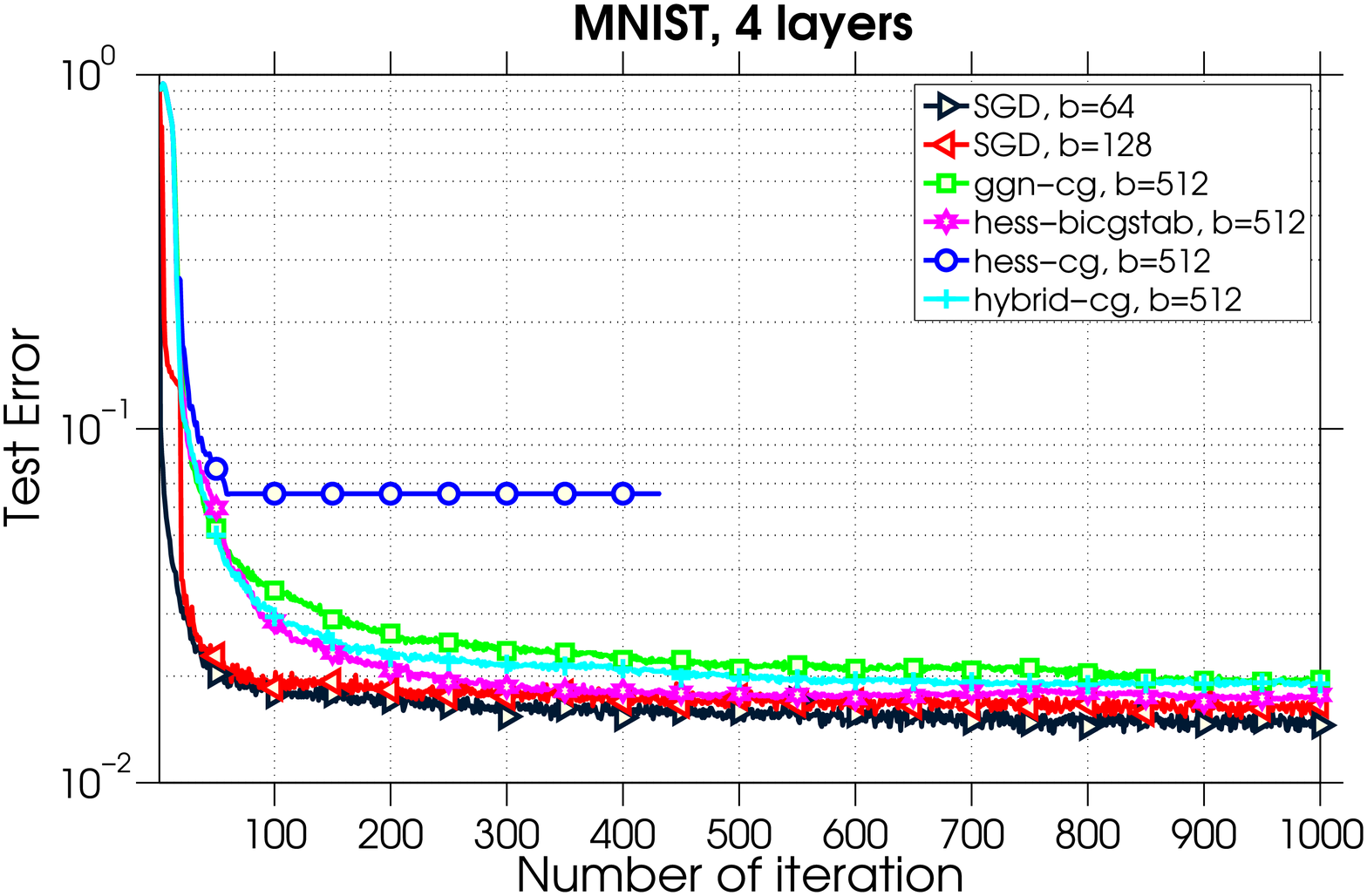}
\includegraphics[width=4cm]{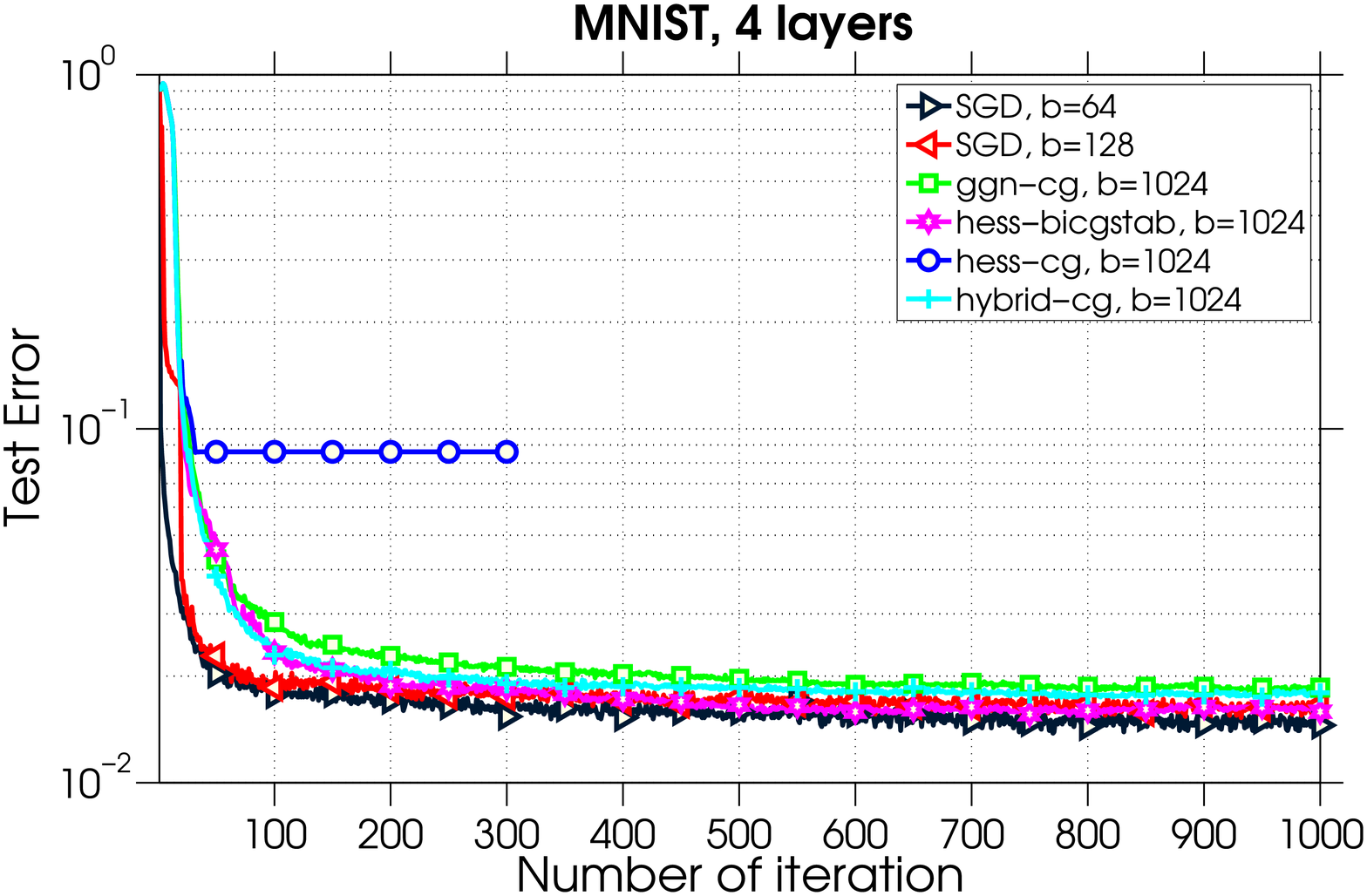}
\includegraphics[width=4cm]{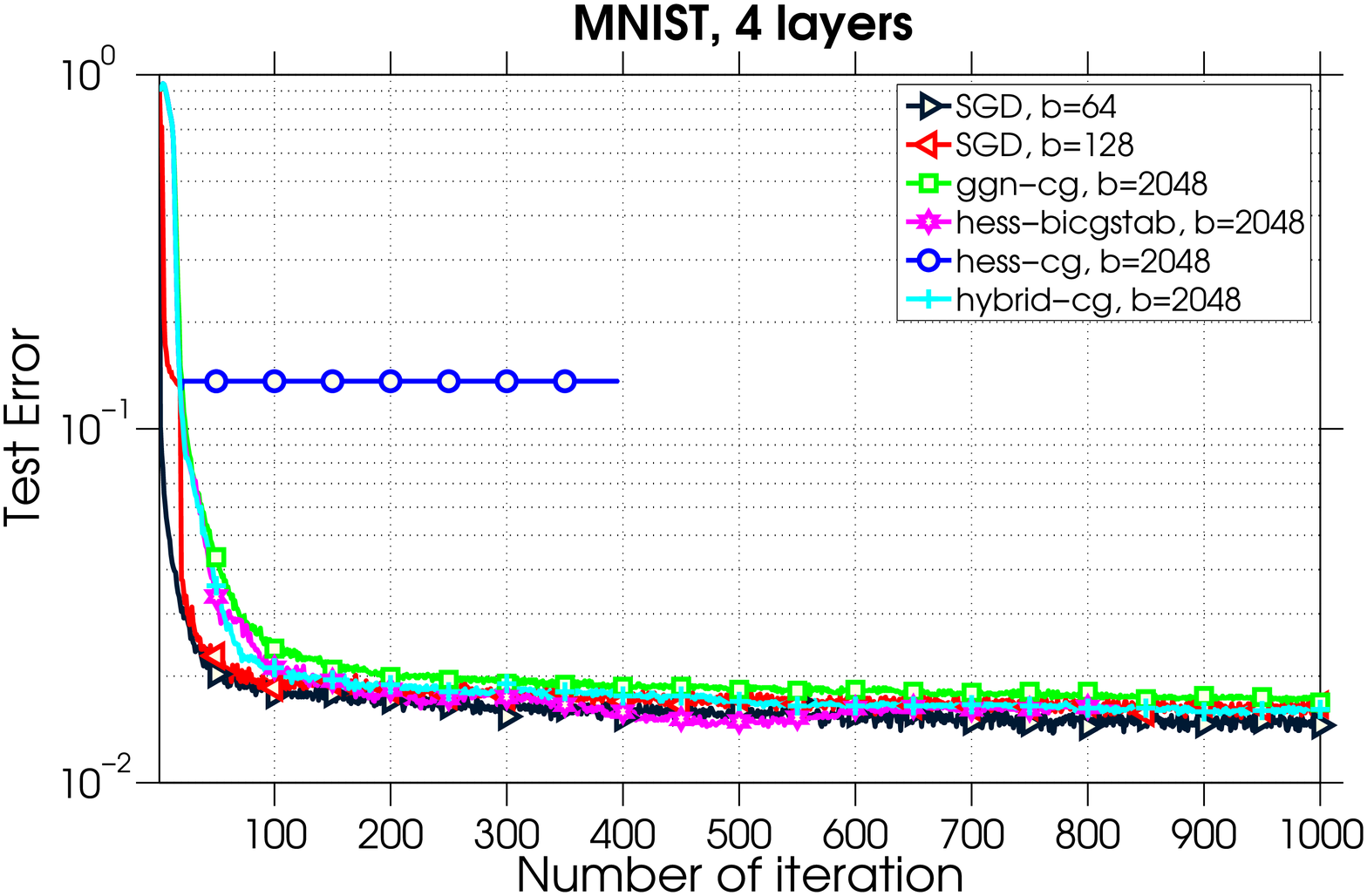}
\caption{Performance comparison among various size of mini-batches on different methods (first 3 plots concern training error and last 3 plots concern testing error).
and number of iterations required to obtain training error 0.02 as a function of batch size for second order methods.
The neural network has two hidden layers with size
400, 150.
}
\label{fig: comparison_four_layers}
\end{figure*}

We train MNIST (images) and TIMIT (speech) dataset with various number of hidden layers and hidden units. Note that we do not do any distortions or pre-training for these two dataset as we are interested in scaling and stability of the methods.

{\bf Comparison of Distributed SGD and Distributed Hessian-free Variants.}
In Figure \ref{fig: comparison_sgd_hf_performance}

 We train MNIST dataset with one hidden layers of $400$ units, with $N=16$ MPI processes and compare the performance of four algorithms in terms of the objective value vs.
 iterations (left), effective passes over data -- epochs (middle) and number of communications (right).  Note that for presentation purposes we count one epoch of SGD as "one iteration", even-though  it is $n/(N\times b)$ iterations.
If we look on the evolution of objective value vs. iterations, all algorithms looks very comparable, however, if we check the evolution of objective value vs. epochs, we see that each iteration of second order method requires multiple epochs (one epoch for computing full gradient and possibly many more for a line-search procedure). This can be seen as the trade-off due larger mini-batch sizes, because of which the number of updates within an epoch (one-pass through all the samples) is reduced. We currently looking into methods to address this issue which typical of large-batch second order methods.
We would like to stress, that in a contemporary high performance clusters each node is usually massively parallel (e.g. in our case 2.65 Tflops) and communication is usually a bottleneck.
The very last plot in Figure \ref{fig: comparison_sgd_hf_performance} shows the evolution of objective value with respect to communication. As it is apparent, SGD needs in order of magnitude more communication (for 1 epoch it needs $n/(Nb)$ communications).
However, increasing $b$ would decrease number of communications per epoch, but it would significantly decrease the convergence speed. We can also see that SGD got stuck around training error 0.01, whereas second order methods continues to make significant additional progress.

In Figure \ref{fig: comparison_four_layers} we show how  increasing the size of a batch is accelerating convergence of second order methods.
On contrary, increasing batch size for SGD from $b=64$ to $b=128$ (beyond which the SGD-performance largely deteriorates). This also implies that increasing batch size to decrease communication overhead of SGD will slow down the method.
Hybrid-CG is a method that  uses Hessian information and Gauss-Newton information alternatively. At the beginning, when the starting point may be far away from local minimizer, we use HF-CG method and whenever a negative curvature is encountered, we turn to use Gauss-Newton Hessian approximation for next iteration, and after this iteration,  HF-CG is used again.
The intuition behind it is that we want to use the exact Hessian information as much as possible but also expected to have a valid descent direction at each iteration.
From Figure \ref{fig: comparison_four_layers}, we observe that unlike SGD method, Hessian-free variants (except HF-CG), are able to make further progress by reducing objective value of error functions, as well as training error continuously.
Meanwhile, our proposed HF-Bi-CG-STAB outperforms other Hessian-free variants, which shows consistently in all three figures (and others figures in Appendix).
If we consider the scaling property in terms of mini-batch, we can see that as the size of mini-batch increase, Hessian-free variants actually {\it performs better}. The intuition behind it is that larger $b$ is making the stochastic Hessian approximation much closer to the true Hessian.
Figure \ref{fig: comparison_four_layers} right
shows scaling of convergence rate as a function of  mini-batch.
In the plot, $b$ represents the size of mini-batch and the $y$-axis is the number of iteration the algorithm needed to hit training error $0.04$. We see that as we increase the size of mini-batches, it takes less iteration to achieve a training error threshold. The reason is that with a larger mini-batches, we are able to approximate the Hessian more accurate and it is then good to find an aggressive descent direction.

{\bf Scaling Properties of Distributed Hessian-free Methods.}
Let us now study scaling properties of existing and proposed distributed Hessian-free methods.
All experiments in this section were done on the large TIMIT speech recognition data-set, with $360$ features, $1973$ classes, and $1013950$ samples.
 The samples are split into two parts, where we use $70\%$ as training data-set and $30\%$ as testing data-set.
The network is set to have $3$ fully-connected hidden layers with $512$ units each.
In Figure \ref{fig: Timit_all_appendix} (top-left)
we show the scaling or all studied second order methods with respect to the number of nodes. Each node has two sockets, which correspond to two
 non-uniform memory ({\sc NUMA}) regions. To exploit this we run a MPI rank per socket and within the socket we use the multi-threaded Intel MKL functions for the BLAS kernels (sgemm, sgemv), which make up the core compute - to utilize the available $18$ cores.\\

The picture on left shows how the duration of one iteration scale with number of nodes for various size of batch size.
Observe, that the scaling is almost linear for values $B\geq 4096$.
Actually, the small batch size is the primary bottleneck for scaling because of the limited parallelism. Hence this larger batch-size (increased parallelism) is essential for scaling to larger number of nodes.
As was show in \ref{fig: comparison_four_layers}
large batch-size are generally only beneficial for second order methods (as opposed to SGD).
Figure \ref{fig: Timit_all_appendix} (top, last 3 plots)
shows the speed-up property of the 3 main components of the second order algorithm.
Note that both gradient computation and line search inherit similar behavior as the total cost of one iteration.
In case of CG, we see that the time of one CG is increasing with increasing size of nodes.
The reason for it is that Hessian-vector product is evaluated only for one batch (whose time should be independent from the number of nodes used)
but the communication time is naturally increased with mode nodes.
It reminds us to remark that the time of communication in this case is comparable to the local compute and hence the pictures suggest very bad scaling. Let us stress that the time of one CG is in order of magnitude smaller then computing of full gradient or line search procedure. As an immediate next step, we are looking into more comprehensive characterization of the compute and bottleneck analysis of both single and multi-node performance.
Figure \ref{fig: Timit_all_appendix} (bottom) shows the each batch size the time of 3 major components of the algorithm.

\section{Conclusion}

In this paper, we revisited HF optimization for deep neural network, proposed a distributed variant with analysis. We showed that unlike the parallelism of SGD, which is inherently sequential, and has limitation (large batch-size helps to scale it but slows convergence). Moreover, a cheap way to detect curvature information and use negative curvature direction   by using BI-CG-STAB method is discussed. It is known that to use of negative curvature direction is essential on improves the training performance.
Furthermore, a Hybrid variant is discussed and applied. We show a significant speed-up by applying distributed HF in numerical experiment and the basic comparison among SGD and other HF method shows competitive performance.

\section*{Acknowledgments}
{
This research was supported by National Science Foundation grant (CCF-1618717).
}

%
%
%
%
%

 \small
\bibliographystyle{aaai}
\bibliography{ref}

\end{document}

%% file: everything.tex
\usepackage{amsmath,amssymb, amsthm}

\usepackage{fullpage}

\usepackage{multirow}
\usepackage{amsfonts}
\usepackage{layout}
\usepackage{url}
\usepackage{color}
\usepackage{graphicx}
\usepackage{algorithmic}
\usepackage{algorithm}
\usepackage{verbatim}
\usepackage{epsfig}
\usepackage{tikz}

\PassOptionsToPackage{normalem}{ulem}
\usepackage{ulem}

\usepackage{cases}

\providecolor{added}{rgb}{0,0,1}
\providecolor{deleted}{rgb}{1,0,0}

\definecolor{orange}{RGB}{255,127,0}

\newcommand{\R}{\mathbb{R}}

\newcommand{\mc}[1]{\mathcal{#1}}

\theoremstyle{plain}

\theoremstyle{definition}

\newcommand*{\starnr}{\stepcounter{equation}\tag{\theequation}}

%% file: large_scale_dnn.bbl
\begin{thebibliography}{}

\bibitem[\protect\citeauthoryear{Amodei \bgroup et al\mbox.\egroup
  }{2015}]{deepspeech2}
Amodei, D.; Anubhai, R.; Battenberg, E.; Case, C.; Casper, J.; Catanzaro, B.;
  Chen, J.; Chrzanowski, M.; Coates, A.; Diamos, G.; et~al.
\newblock 2015.
\newblock Deep speech 2: End-to-end speech recognition in english and mandarin.
\newblock {\em arXiv:1512.02595}.

\bibitem[\protect\citeauthoryear{Das \bgroup et al\mbox.\egroup
  }{2016}]{das2016distributed}
Das, D.; Avancha, S.; Mudigere, D.; Vaidynathan, K.; Sridharan, S.; Kalamkar,
  D.; Kaul, B.; and Dubey, P.
\newblock 2016.
\newblock Distributed deep learning using synchronous stochastic gradient
  descent.
\newblock {\em arXiv:1602.06709}.

\bibitem[\protect\citeauthoryear{Dauphin \bgroup et al\mbox.\egroup
  }{2014}]{dauphin2014identifying}
Dauphin, Y.~N.; Pascanu, R.; Gulcehre, C.; Cho, K.; Ganguli, S.; and Bengio, Y.
\newblock 2014.
\newblock Identifying and attacking the saddle point problem in
  high-dimensional non-convex optimization.
\newblock In {\em NIPS},  2933--2941.

\bibitem[\protect\citeauthoryear{He \bgroup et al\mbox.\egroup
  }{2015}]{heResNet}
He, K.; Zhang, X.; Ren, S.; and Sun, J.
\newblock 2015.
\newblock Deep residual learning for image recognition.
\newblock {\em arXiv:1512.03385}.

\bibitem[\protect\citeauthoryear{Hinton \bgroup et al\mbox.\egroup
  }{2012}]{hinton2012deep}
Hinton, G.; Deng, L.; Yu, D.; Dahl, G.~E.; Mohamed, A.-r.; Jaitly, N.; Senior,
  A.; Vanhoucke, V.; Nguyen, P.; Sainath, T.~N.; et~al.
\newblock 2012.
\newblock Deep neural networks for acoustic modeling in speech recognition: The
  shared views of four research groups.
\newblock {\em Signal Processing Magazine, IEEE} 29(6):82--97.

\bibitem[\protect\citeauthoryear{Kiros}{2013}]{kiros2013training}
Kiros, R.
\newblock 2013.
\newblock Training neural networks with stochastic hessian-free optimization.
\newblock {\em arXiv:1301.3641}.

\bibitem[\protect\citeauthoryear{Krizhevsky, Sutskever, and
  Hinton}{2012}]{krizhevsky2012imagenet}
Krizhevsky, A.; Sutskever, I.; and Hinton, G.~E.
\newblock 2012.
\newblock Imagenet classification with deep convolutional neural networks.
\newblock In {\em Advances in neural information processing systems},
  1097--1105.

\bibitem[\protect\citeauthoryear{Martens}{2010}]{martens2010deep}
Martens, J.
\newblock 2010.
\newblock Deep learning via hessian-free optimization.
\newblock In {\em ICML},  735--742.

\bibitem[\protect\citeauthoryear{Nocedal and
  Wright}{2006}]{nocedal2006numerical}
Nocedal, J., and Wright, S.
\newblock 2006.
\newblock {\em Numerical optimization}.
\newblock Springer Science \& Business Media.

\bibitem[\protect\citeauthoryear{Olivares, Moguerza, and
  Prieto}{2008}]{olivares2008nonconvex}
Olivares, A.; Moguerza, J.~M.; and Prieto, F.~J.
\newblock 2008.
\newblock Nonconvex optimization using negative curvature within a modified
  linesearch.
\newblock {\em European Journal of Operational Research} 189(3):706--722.

\bibitem[\protect\citeauthoryear{Polyak and
  Juditsky}{1992}]{polyak1992acceleration}
Polyak, B.~T., and Juditsky, A.~B.
\newblock 1992.
\newblock Acceleration of stochastic approximation by averaging.
\newblock {\em SIAM Journal on Control and Optimization} 30(4):838--855.

\bibitem[\protect\citeauthoryear{Saad}{2003}]{saad2003iterative}
Saad, Y.
\newblock 2003.
\newblock {\em Iterative methods for sparse linear systems}.
\newblock Siam.

\bibitem[\protect\citeauthoryear{Seide \bgroup et al\mbox.\egroup
  }{2014}]{seide2014parallelizability}
Seide, F.; Fu, H.; Droppo, J.; Li, G.; and Yu, D.
\newblock 2014.
\newblock On parallelizability of stochastic gradient descent for speech dnns.
\newblock In {\em Acoustics, Speech and Signal Processing (ICASSP), 2014 IEEE
  International Conference on},  235--239.
\newblock IEEE.

\bibitem[\protect\citeauthoryear{Simonyan and
  Zisserman}{2014}]{simonyan2014very}
Simonyan, K., and Zisserman, A.
\newblock 2014.
\newblock Very deep convolutional networks for large-scale image recognition.
\newblock {\em arXiv:1409.1556}.

\bibitem[\protect\citeauthoryear{Sutskever \bgroup et al\mbox.\egroup
  }{2013}]{sutskever2013importance}
Sutskever, I.; Martens, J.; Dahl, G.; and Hinton, G.
\newblock 2013.
\newblock On the importance of initialization and momentum in deep learning.
\newblock In {\em ICML},  1139--1147.

\bibitem[\protect\citeauthoryear{Tak{\'a}{\v{c}} \bgroup et al\mbox.\egroup
  }{2013}]{takavc2013mini}
Tak{\'a}{\v{c}}, M.; Bijral, A.; Richt{\'a}rik, P.; and Srebro, N.
\newblock 2013.
\newblock Mini-batch primal and dual methods for svms.
\newblock In {\em ICML}.

\bibitem[\protect\citeauthoryear{Wiesler, Li, and
  Xue}{2013}]{wiesler2013investigations}
Wiesler, S.; Li, J.; and Xue, J.
\newblock 2013.
\newblock Investigations on hessian-free optimization for cross-entropy
  training of deep neural networks.
\newblock In {\em INTERSPEECH},  3317--3321.

\bibitem[\protect\citeauthoryear{Zhang}{2016}]{zhang2016distributed}
Zhang, S.
\newblock 2016.
\newblock {\em Distributed stochastic optimization for deep learning}.
\newblock Ph.D. Dissertation, New York University.

\end{thebibliography}
